# Framework for Collaborative Operation of Autonomous Delivery Vehicles Within a Marshaling Yard

James O'Hara
*Autonomous Systems Research Center*
Noblis Inc
Reston, VA USA
james.ohara@noblis.org

Karl Wunderlich
*Autonomous Systems Research Center*
Noblis Inc
Reston, VA USA
karl.wunderlich@noblis.org

Gregory Stevens
*M-City*
University Of Michigan
Ann Arbor, MI USA
gregstvn@umich.edu

*Abstract*— As autonomous vehicles slowly deploy into urban roads for limited use cases with significant edge case issues, closed facilities like marshaling yards provide a ripe case for combining lower-level vehicle autonomy with fixed infrastructure to create full autonomy without similar edge case concerns. Within a delivery marshaling yard, electric fleet vehicles complete a set of sequential tasks (charging, inspection, cleaning, and loading) before exiting the yard with their new load of deliveries. Hybrid automation of the vehicles and infrastructure can allow these vehicles to reach full autonomy and navigate the facility without the need of a driver, allowing for quicker movement between tasks increasing vehicle throughput. However, isolated autonomous operations based on static rules are prone to gridlock causing facility failures that temporarily shut down operations. Our orchestrated autonomy solution uses decentralized, dynamic priority scoring of vehicles based on the current status of the marshaling yard to optimally assign vehicles to tasks to increase vehicle throughput. Using a simulated facility with three marshaling yard sizes (small, medium, and large) and three demand levels (low, medium, high), we demonstrated that our orchestration solution increases vehicle throughput above static, isolated autonomy for all combinations of yard size and demand, while reducing facility failures at high demand levels.

*Keywords—autonomous vehicles, marshaling yard, delivery vehicles, orchestration, collaborative autonomy,*

## I. Introduction

Depot operations for fleet vehicles include a range of sequential activities, including vehicle loading and unloading, charging, safety inspection and maintenance, cleaning, and dispatching. These operations are conducted sequentially throughout the workday but are subject to peak demand periods where valuable driver/operator and vehicle time is wasted queueing at activity stations. Further, queued vehicles obstruct the movement of depot staff and vehicles-in-motion, resulting in both negative safety and efficiency conditions. Under exceptional demand conditions, queued vehicles can produce gridlock conditions with the facility dramatically reducing depot productivity.

Under Industry 4.0, autonomous vehicles will take on many of the navigation tasks currently handled by delivery drivers, however prior research has focused mainly on the dispatching and routing of delivery vehicles outside of the marshaling yard. Within the delivery vehicle marshaling yard, research focuses on the automation of package handling and load inside the warehouse instead of the operations outside of it. When outside-the-warehouse operations are considered, it is at the larger scale of shipyards or train yards, where the tasks being performed by the autonomous vehicles are closer to those inside a delivery warehouse.

Many autonomous vehicles together in and around a marshaling yard can quickly degenerate into gridlock conditions when each vehicle is operating independently under isolated autonomous control. Our framework can resolve these problems through collective vehicle prioritization, task allocation, and path planning, reducing the rate of gridlock conditions and increasing the throughput of vehicles exiting the facility with new deliveries.

## II. Related work

We were unable to identify specific work related to autonomous delivery vehicle operations within a marshaling yard. However, there are similar use cases with associated research into finding optimal solutions for task assignments to a group of autonomous vehicles. These cases present solutions that can inform the optimal solution to the delivery marshaling yard problem, however key differences in problem formulation mean these solutions cannot be directly applied to this use case.

### A. Warehouse Operations

Autonomous warehouse operations encompass a wide variety of tasks from e-commerce to manufacturing. In these cases, a team of autonomous vehicles within a closed system must accomplish a series of tasks in the quickest amount of time possible to increase production. Multiple approaches to optimal organizing of autonomous vehicles within a warehouse have been researched including cellular warehousing [1], Particle Swarm Optimization [2], and Context-Aware Cloud Robotics [3]. In each case, a critical element in solving the optimization problem was including information from the infrastructure to understand key context in scheduling and assigning tasks. Decentralized control was better able to handle large systems with more vehicles as local and global decision making was able to be decoupled [4]. However, unlike the delivery marshaling yard these techniques optimize for task throughput rather than vehicle throughput as the vehicles are never expected to leave the warehouse.

*B. Non-delivery Marshaling Yards*

Research into autonomous marshaling yards focuses more on rail yards and shipyards where operations are more closely akin to the warehouse than a delivery vehicle marshaling yard. In these use cases, autonomous vehicles are used for maneuvering and reconfiguring shipping containers between different areas of the yard to support freight operations. Some research has focused on changing yard design and management to take more advantage of the different capabilities of autonomous vehicles compared to human operators[5]. Like in warehouse operations, dynamic scheduling that includes contextual information about the tasks and infrastructure are able to improve on static rules-based systems [6]. The solutions identified in [5] and [6] focus their optimization on task throughput instead of vehicle throughput as these operations use specialized vehicles that remain in the yard at all time similar to a warehouse.

*C. Autonomous Dispatching*

Autonomous dispatching presents a similar challenge in optimal organization of autonomous vehicles across one or multiple tasks. For autonomous vehicle parking assignment, a distributed algorithm was found to produce high quality solutions that can be calculated in short amounts of time even under high demand [7]. When multiple tasks were considered for a mobility-on-demand system, unique assignment algorithms were used for each task to calculate the optimal edges between two bipartite graphs of vehicles and tasks which fed into a neural network to decide the ultimate solution [8]. Like the delivery marshaling yard, these solutions focused on optimizing vehicle throughput, but they were operating within a full urban area and did not have to deal with the same constraints on resources and space.

## III. CONCEPT AND SYSTEM ARCHITECTURE

Our decentralized orchestrated autonomy framework utilizes priority scoring to determine the optimal edges of a bipartite graph of vehicles and stations with available capacity. The priority scoring incorporates dynamic contextual information from both the vehicles and infrastructure including the battery level of the vehicle, the remaining tasks the vehicle has to complete, the time since entrance, and a trust score. Equation (1) shows how these factors are combined into a single priority score p for a given vehicle, v. Where $b_v$ is the inverse of the amount of time needed to charge the battery of v to 100 percent. $c_v$ is the number of stations already completed by v, $t_v$ is the current time in seconds for vehicle v above the mean expected circuit completion time, and $\tau_v$ is the trust score for vehicle v.

$$p_v = (60 * b_v) + (20 * c_v) + (5 * t_v) + \tau_v \quad (1)$$

As charging is the station that will take the longest amount of time (DC fast charging of a Rivian electric van from 0-100 percent is two hours), higher priority is placed on vehicles with greater battery levels that will require less charging. Vehicles that are closer to completing the station circuit are given higher priority as they will be able to exit the facility sooner. To prevent vehicles with a lower initial priority from getting stuck in the parking lot, additional priority is given to vehicles that have been in the facility for longer than the expected minimum time to complete the circuit. The trust score is based on the past performance of the vehicle based on peer vehicle ratings in cooperative path planning and efficient operations and is meant to account for heterogeneous behaviors in autonomous vehicles due to hardware, software, or other manufacturer and operator differences.

Re-planning is performed in a decentralized manner without communicated consensus to reduce bandwidth usage. Each vehicle determines its optimal assignment by calculating the assignments for all vehicles currently in the marshaling yard. Priority scores are calculated by all vehicles not currently completing a task whenever a new vehicle enters the marshaling yard or a vehicle completes its time at a station. Re-planning is then performed for each vehicle with a priority score below the new vehicle to determine the optimal station assignment. Each vehicle is considered in order from highest priority score to lowest following these steps:

*1) Check if the vehicle had a previous station assignment and that station has capacity:* If yes, the vehicle maintains that assignment, otherwise the steps are continued

*2) For each station the vehicle still has to visit check if it has met the prerequisites for that station and if the station has capacity*

*3) For each station that fulfills these requirments, find the station closest to the vehicle's current location and assign the vehicle to that station*

*4) If no station meets these requirements, assign the vehicle to the parking lot or other available space*

The vehicles then determine their optimal route to the station and broadcast it to the group. If two or more vehicles intend to use the same space at the same time, they use the priority score to determine which vehicle should be given priority in the space and the other vehicle then re-computes a non-conflicting path.

To maintain situational awareness and support priority score calculations, each vehicle broadcasts its current position, future path, battery charge level, and remaining stations. Each station is set up with a radio to broadcast current utilization.

## IV. EXPERIMENTAL DESIGN

To verify the effectiveness of priority scoring in increasing vehicle throughput in a marshaling yard, we created a simulation environment based on the general delivery marshaling yard layout (Fig 1).

To isolate the impacts of the priority scoring, we simplified the simulation to remove the broadcast component and instead used an open-source Space-Time A* [9] to assign non-conflicting paths in priority order. Additionally, advanced features like utilizing open space as temporary parking or advanced scheduling ahead of task completion were not included. As there is no prior performance to reference, trust scores are set at random on vehicle entrance.

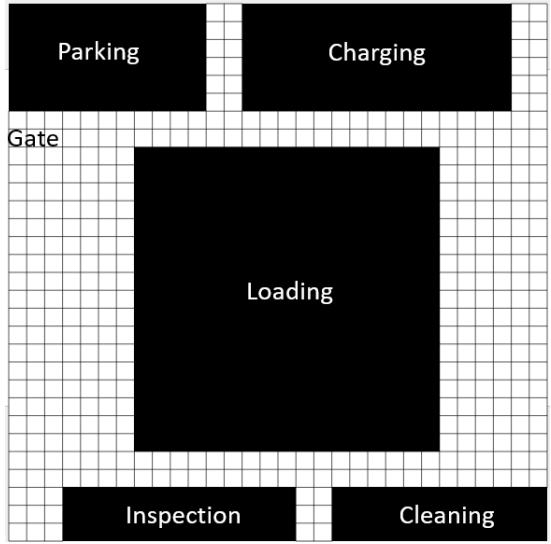

Fig. 1. Scaled simulation marshaling yard setup. (*Each grid square represents 10 meters by 10 meters, black space is not traversable.*)

The simulation environment is meant to represent a 5 hour peak activity period with vehicles potentially entering the yard at a set rate of time depending on the simulated demand and following a Poisson distribution to represent the random clustering of vehicles.

Within the marshaling yard, vehicles were required to visit the charging, inspection, cleaning, and loading stations prior to exiting the facility. Loading was the final task and could not be executed prior to completing the charging, inspection, and cleaning tasks. Time to task completion was simulated as normal distributions as defined in Table 1. Charging time was assigned at vehicle entrance into the network to represent the remaining charge in the vehicle's battery on entrance. A random rate of 0.5 percent of vehicles failed inspection and reduced the parking capacity by one vehicle for the rest of the simulation run. Vehicles were set to travel at a consistent 16.1 kilometers per hour (10 miles per hour).

TABLE I. STATION TASK COMPLETION DISTRIBUTIONS

| Station | Task Completion Normal Distributions | |
|---|---|---|
| | *Mean* | *Standard Deviation* |
| Charging | 60 minutes | 30 minutes |
| Cleaning | 20 minutes | 2 minutes |
| Inspection | 10 minutes | 2 minutes |
| Loading | 20 minutes | 2 minutes |
| Parking | 2 minutes | 2 minutes |

Two operational conditions were defined: isolated autonomous vehicles and our orchestrated framework. For the isolated operational condition, vehicles behaved according to static rules and followed defined paths between each station in a loop attempting to complete each task. Under this condition, no dynamic context is provided to the vehicle for planning. A vehicle does not know the capacity status of each station until it arrives at the entrance gate. If the station has capacity, the vehicle enters the station and begins the task, otherwise the vehicle continues driving to the next station it needs to complete. If all required stations for a given vehicle are full, that vehicle will attempt to enter the parking area and wait to try again.

Under both operational conditions, a facility failure occurred when a vehicle found all stations and the parking lot at capacity, leaving it stranded in the roadway network. At this time the simulation run was stopped and marked as a facility failure.

The two operational conditions were tested with three yard sizes as defined by the station capacities shown in Table 2.

TABLE II. STATION CAPACITIES BY YARD SIZE

| Stations | Yard Sizes | | |
|---|---|---|---|
| | *Small* | *Medium* | *Large* |
| Charging | 14 vehicles | 28 vehicles | 42 vehicles |
| Cleaning | 10 vehicles | 20 vehicles | 40 vehicles |
| Inspection | 10 vehicles | 20 vehicles | 40 vehicles |
| Loading | 16 vehicles | 30 vehicles | 68 vehicles |
| Parking | 30 vehicles | 60 vehicles | 90 vehicles |

For each yard size, at least three demand levels were tested, to represent high, medium, and low demand. For the small yard, we tested 60, 80, and 100 vehicles. For the medium yard, we tested 80, 160, and 225 vehicles. For the large yard, we tested 160, 225, and 340 vehicles.

V. RESULTS

For all yard size and demand level combinations, the orchestrated system had a higher vehicle throughput and lower facility failure rate at high demand levels. Even without taking advantage of more advanced features of orchestration like temporary parking or advanced scheduling, the increase in throughput from priority scoring alone was enough to reduce gridlock scenarios.

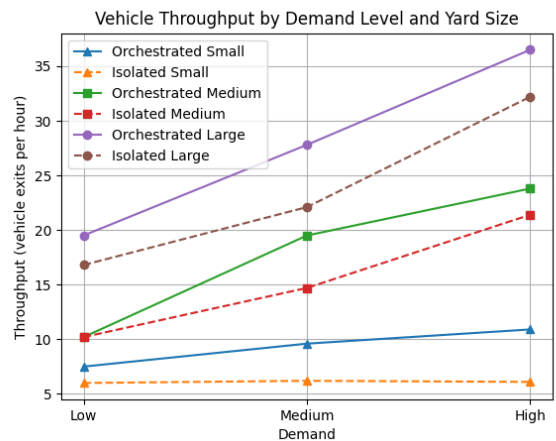

Fig. 2. Vehicle Throughput for Orchestrated and Isolated Autonomy by Demand Level and Yard Size.

The orchestrated vehicles were able to successfully use the dynamic context, prioritization scoring, and bipartite task assignment to reduce wasted movement and more optimally use limited station capacity (Fig. 2). In successful runs, vehicle throughput was increased on average by 3.4 vehicles per hour across all demand levels and yard sizes.

As the small and medium yard size results show, the benefits of orchestrated as compared to isolated autonomy were reduced as the demand increased beyond the abilities of the orchestration to maneuver the vehicles through the system. While orchestration allows for more optimal station assignments the stations themselves still present a bottleneck with their limited capacities and time to complete that will eventually cause a facility failure when demand is too high.

While increasing demand will eventually lead to facility failure, the increased throughput of the orchestrated framework allowed it to handle the increased demand better than the isolated autonomy. While the isolated scenarios saw a rapid increase in facility failures at high demand levels (mean 33.3 percentage point increase), the orchestrated framework was able to flatten the curve for a more modest increase of 4.7 percentage points (Fig. 3).

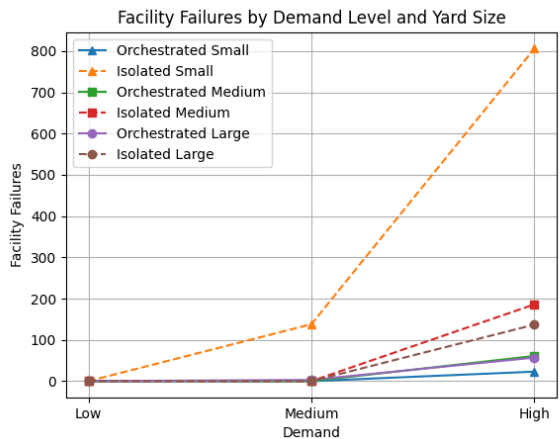

Fig. 3. Number of Facility Failures for Orchestated and Isolated Autonomy by Demand Level and Yard Size.

## VI. Conclusion

Orchestration of autonomous delivery vehicles within a marshaling yard can increase vehicle throughput and reduce potential facility failures due to gridlock compared to isolated, static autonomous operations. Decentralized task assignment and coordination of movement through priority scoring allows for smoother operations, especially when the yard size is smaller and vehicles quickly fight for space and overwhelm station capacities in isolated operations.

Our orchestrated framework addresses a research gap in optimizing task assignment for vehicle throughput rather than task throughput in use cases like the marshaling yard. Our framework was able to successfully increase vehicle throughput and reduce facility failures across all combinations of demand level and yard size.

Additional research should be focused on adding additional advanced functionality, like utilizing all open space within the facility as temporary parking areas, which would reduce the need for dedicated parking and open up more space for additional station capacity. Further research should also focus on identifying a closed-form equation for identifying the optimal ratio of yard size to vehicle demand for vehicle throughput.

In the long term, a fully autonomous delivery marshaling yard could incorporate decentralized coordination between the delivery vehicles and warehouse robotics to do more dynamic loading operations that similarly open space for more flexible usage and increase vehicle throughput.